\documentclass[10pt,twocolumn,letterpaper]{article}

\usepackage{iccv}
\usepackage{cite}
\usepackage{times}
\usepackage{epsfig}
\usepackage{graphicx}
\usepackage{amsmath}
\usepackage{amssymb}
\usepackage{multirow}
\usepackage{longtable}
\usepackage{color}
\usepackage{rotating}
\usepackage{makecell}
\usepackage{array}
\usepackage{booktabs}
\usepackage{colortbl}
\usepackage{arydshln}
\usepackage{mathrsfs}
\usepackage{amsmath}
\usepackage{url}

\usepackage[breaklinks=true,bookmarks=false]{hyperref}

\iccvfinalcopy 

\def\iccvPaperID{****} 

\DeclareRobustCommand*{\IEEEauthorrefmark}[1]{%
\raisebox{0pt}[0pt][0pt]{\textsuperscript{\footnotesize\ensuremath{#1}}}}
\begin{document}

\title{\LaTeX\ Author Guidelines for ICCV Proceedings}

\title{More About Covariance Descriptors for Image Set Coding: Log-Euclidean Framework based Kernel Matrix Representation}

\author{Kai-Xuan Chen\textsuperscript{1} \and Xiao-Jun Wu\textsuperscript{1,*}\and Jie-Yi Ren\textsuperscript{1,2} \and Rui Wang\textsuperscript{1} \and Josef Kittler\textsuperscript{2}\and
\IEEEauthorrefmark{1}School of IoT Engineering, Jiangnan University,Wuxi, 214122, China.\and
\IEEEauthorrefmark{2}Center for Vision, Speech and Signal Processing, University of Surry, GU2 7XH, Guildford, UK.\\
{\tt\small kaixuan\_chen\_jnu@163.com wu\_xiaojun@jiangnan.edu.cn jieyi.ren@surrey.ac.uk}\\
{\tt\small cs\_wr@jiangnan.edu.cn j.kittler@surrey.ac.uk}}

\maketitle

\begin{abstract}
	\indent We consider a family of structural descriptors for visual data, namely covariance descriptors (CovDs) that lie on a non-linear symmetric positive definite (SPD) manifold, a special type of Riemannian manifolds. We propose an improved version of CovDs for image set coding by extending the traditional CovDs from Euclidean space to the SPD manifold. Specifically, the manifold of SPD matrices is a complete inner product space with the operations of logarithmic multiplication and scalar logarithmic multiplication defined in the Log-Euclidean framework. In this framework, we characterise covariance structure in terms of the arc-cosine kernel which satisfies Mercer's condition and propose the operation of mean centralization on SPD matrices. Furthermore, we combine arc-cosine kernels of different orders using mixing parameters learnt by kernel alignment in a supervised manner. Our proposed framework provides a lower-dimensional and more discriminative data representation for the task of image set classification. The experimental results demonstrate its superior performance, measured in terms of recognition accuracy, as compared with the state-of-the-art methods. 
\end{abstract}
\section{Introduction}
	\indent The representation of visual data plays a vital role in identifying the content of images \cite{perronnin2010improving,jegou2012aggregating}, image sets \cite{tan2015grassmann,chen2018riemannian} and videos \cite{peng2014action,feng2017object}.  There are many descriptors for visual data. As is well known, the most popular representations for recognition tasks are bag-of-visual-words (BoVW) models \cite{sivic2003video}, fisher vectors (FV) \cite{jegou2012aggregating} and vector of locally aggregated descriptors (VLAD) \cite{arandjelovic2013all}. These representations are ultimately in the form of vectors and typically involve the following four steps: extracting features, generating codebook, encoding and pooling, and normalization.\\
	\indent Structured representations, such as linear subspaces \cite{hamm2009extended}, and covariance descriptors (CovDs) \cite{tuzel2008pedestrian,wang2012covariance}, have recently been shown to offer efficient and powerful representations for high dimensional tasks in computer vision. In particular, CovDs, defined by the second-order statistics of sample features, have been widely used as the visual representations for both single image \cite{tuzel2008pedestrian} and image sets \cite{wang2012covariance}. Prior to CovDs for describing image sets studied in \cite{wang2012covariance}, covariance matrices have been used as region covariance descriptors to characterise  local regions within an image. They have been  applied to the task of object tracking, object detection and texture classification. In contrast, when CovDs are used to describe image sets \cite{wang2012covariance}, samples are the images in the set and features are the raw intensities of the image pixels. The resulting covariance matrices are often singular because the feature dimensionality (the number of the pixels in the image) is usually larger than the number of samples (the number of images). Among the aforementioned methods, CovDs for describing image sets has the following characteristics: \\	
	\indent 1) In contrast to the vector representations of BoVW, FV and VLAD, which generate linear descriptors via codebooks, CovDs directly generate structured representations.\\
	\indent 2) A feature matrix of an image set with $n$ images: $S$ = [$s_1$,$s_2$,$...$,$s_n$], where $s_i$ $\in$ $R^d$ is a $d$-dimensional feature vector characterising the $i$-th image. Here, $d$ is the number of the pixels in each image. \\
	\indent 3) The resulting covariance matrix $C$ $\in$ $R$$^{d \times d}$ tends to be singular and high dimensionality, and may contain a certain amount of redundant information.\\
	\indent Characteristic 3 summarises several deficiencies of traditional CovDs as image set descriptors. In this paper, we propose a improved framework,  which involves using a kernel matrix defined in terms of representations associated with sub-image sets, instead of pixels. Our proposed framework enables to generate covariance descriptors on non-linear SPD manifold (CovDs-S\footnote{Source code: \url{https://github.com/Kai-Xuan/iCovDs}}). The experimental results show the advantages of our proposed CovDs-S.\\
	\indent The rest of this paper is organized as follows: In Section 2, we introduce the theory of Riemannian geometry of SPD manifold and review the Log-Euclidean framework which is the baseline for our proposed approach. In Section 3, we give a brief overview of the traditional CovDs as image set descriptors and present the proposed framework. We present and discuss the experimental results in Section 4.  Section 5 draws conclusions and outlines future work.
\section{Background Theory}
	\indent This section first provides a brief introduction to Riemannian geometry of SPD manifold,  and proposes the process of SPD mean centralization. We then present a Log-Euclidean framework based arc-cosine (LogE.Arc) kernel.\\
	\indent \begin{bfseries} Notation: \end{bfseries} In this paper, $I_n$ is an ${n \times n}$ identity matrix. $S_n^{++}$ denotes the SPD manifold spanned by real ${n \times n}$ SPD matrices and $S_n$ denotes the space spanned by real ${n \times n}$ symmetric matrices. $T_P$$S_n^{++}$ is the tangent space at the point $P$ $\in$ $S_n^{++}$, which is a flat surface spanned by real ${n \times n}$ symmetric matrices. Diag($e_1,e_2,...,e_n$) is a diagonal matrix with the diagonal elements $e_1,e_2,...,e_n$. The matrix logarithm, log($\cdot$): $S_n^{++}$$\to$$S_n$ is defined as:
\begin{equation}
\label{eq1}
\mathop {\log}(X)=U{\rm Diag}(({\log}(e_1,e_2,...,e_n))U^T
\end{equation} 
with $X$ = $U$Diag($e_1,e_2,...,e_n$)$U^T$. If $X$$\in$$S_n^{++}$ is an SPD matrix, log($X$) $\in$ $T_I$$S_n^{++}$ will be a point in the tangent space at the identity matrix $I_n$.  Similarly, the matrix exponential exp($\cdot$): $S_n$$\to$$S_n^{++}$ is defined as:
\begin{equation}
\label{eq2}
\mathop {\exp}(X)=U{\rm Diag}(({\exp}(e_1,e_2,...,e_n))U^T
\end{equation} 
with $X$ = $U$Diag($e_1,e_2,...,e_n$)$U^T$. 
\subsection{The General Metrices on SPD Manifold}
	\indent A real $n \times n$ SPD matrix $X$ $\in$ $S_n^{++}$ satisfies $v^TXv$$\ge$0 for all non-zero $v$ $\in$ $R^n$. The Affine Invariant Riemannian Metric (AIRM) is the frequently studied Riemannian metric on the SPD manifold \cite{pennec2006riemannian}. Beside AIRM, Log-Euclidean Metric (LEM) \cite{arsigny2007geometric} and two types of Bregman divergence \cite{kulis2009low}, namely Stein \cite{sra2012new} and Jeffrey \cite{harandi2014bregman} divergence, are also widely used to analyze SPD matrices.\\
\begin{bfseries} Definition 1 \end{bfseries} (Affine Invariant Riemannian Metric, AIRM) \emph{ The most common Riemannian metric on  $S_n^{++}$ is Affine Invariant Riemannian Metric (AIRM) \cite{pennec2006riemannian}, in which the geodesic distance $d_G$: $S_n^{++}$$\times$$S_n^{++}$$\to$[0,$\infty$) between two SPD matrices $X$ and $Y$ can be obtained by:}
\begin{equation}
\label{eq3}
\mathop {{d}{_{G}^2}}(X,Y) = {\parallel{\log\left( X^{-\frac{1}{2}}YX^{-\frac{1}{2}}\right)}\parallel}{_F^2}
\end{equation} 
where $\parallel{\cdot}\parallel_F$ denotes is the Frobenius norm. log($\cdot$) is the matrix principal logarithm. \\
\begin{bfseries} Definition 2 \end{bfseries} (Stein divergence) \emph{The Stein, or S, divergence \cite{sra2012new} $d_S$: $S_n^{++}$$\times$$S_n^{++}$$\to$[0,$\infty$) is a special type of Bregman divergence:}
\begin{equation}
\label{eq4}
\mathop {d{_S^2}}(X,Y) = {\rm log \,det}(\frac{X+Y}{2}) - \frac{1}{2}{\rm log \,det}(XY)
\end{equation} \\
\begin{bfseries} Definition 3 \end{bfseries} (Jeffrey divergence) \emph{The Jeffrey, or, J, divergence \cite{harandi2014bregman} $d_J$: $S_n^{++}$$\times$$S_n^{++}$$\to$[0,$\infty$) is another type of Bregman divergence:}
\begin{equation}
\label{eq5}
\mathop {d{_J^2}}(X,Y) = \frac{1}{2}Tr(X^{-1}Y) + \frac{1}{2}Tr(Y^{-1}X) - n
\end{equation} 
The Stein and Jeffrey divergence are similar to geodesic distance induced by AIRM \cite{pennec2006riemannian}.\\
\begin{bfseries} Definition 4 \end{bfseries} (Log-Euclidean Metric, LEM) \emph{For two SPD matrices $X,Y$ $\in$ $S_n^{++}$, the Log-Euclidean Distance (LED) \cite{wang2012covariance,arsigny2007geometric}, is defined by Frobenius norm in the tangent space at identity matrix  $I_n$:}
\begin{equation}
\label{eq6}
\mathop {{d}{_{LEM}^2}}(X,Y) =   {\parallel{\log\left(X\right)-\log\left(Y\right)}\parallel}{_F^2}
\end{equation} 
Accordingly, the SPD manifold will be reduced to a flat Riemannian space \cite{arsigny2007geometric} while endowed with LEM. 
\begin{figure}
\begin{center}
\includegraphics[width=0.75 \linewidth]{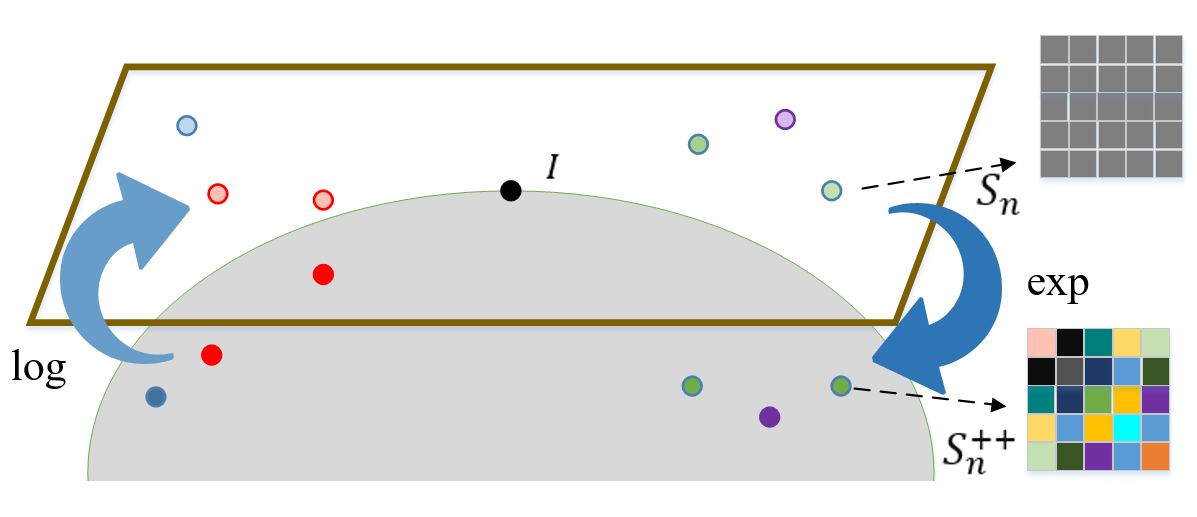} 
\end{center}
   \caption{The illustration of logarithmic and exponential mapping.}
\label{fig:short}
\end{figure}
\subsection{Log-Euclidean Framework}
	\indent In the Log-Euclidean framework, the matrix logarithm log($\cdot$): $S_n^{++}$$\to$$S_n$ is smooth and bijective, and its inverse map, denoted by exp($\cdot$), is smooth as well. Figure 1 illustrates these two operations on SPD manifold. The Log-Euclidean Kernel \cite{wang2012covariance} is derived by computing the inner product in the domain of logarithm matrix:
\begin{equation}
\label{eq7}
\mathop {{k}{_{LogE}}}(X,Y) =  Tr({\log}(X){\log}(Y))
\end{equation} 
where $Tr$ denotes the matrix trace. The Log-Euclidean kernel is a positive definite kernel and has been shown to meet Mercer's conditions in \cite{wang2012covariance}. The operations logarithmic multiplication and scalar logarithmic multiplication are the corresponding Euclidean operations in the domain of logarithm matrix, followed by an inverse mapping back to the SPD manifold via the operation of matrix exponential (Interested readers can refer to \cite{arsigny2007geometric,li2013log} for details). We thus can propose the operation of mean centralization on SPD matrices.  \\
\begin{bfseries} Proposition 1 \end{bfseries} \emph{In line with the brief overview of the Log-Euclidean framework \cite{arsigny2007geometric,li2013log}, we define the operation of mean centralization on SPD matrices in three steps. Firstly, we map the SPD matrices into the domain of logarithm matrix. Then, we centralize the resulting symmetric matrix by an operation that is similar to centering the kernel matrix in \cite{cortes2010two}.  Finally, we map the centralized matrices back to SPD manifold via exponential mapping. For an arbitrary real $n$$\times$$n$ SPD matrix $X$ $\in$ $S_n^{++}$, the operation of mean centralization can be written as:}
\begin{equation}
\begin{aligned}
\label{eq8}
\mathop  {\tilde{X}} = {\rm exp}({\hat{X}}) \qquad \qquad \qquad  \quad  \\
where \quad [{\hat{X}}]{_{i,j}} = [{\log}(X)]{_{i,j}} - \frac{1}{n}\sum_{\substack{i=1}}^n[{\log}(X)]_{i,j} \\
 - \frac{1}{n}\sum_{\substack{j=1}}^n[{\log}(X)]_{i,j} + \frac{1}{n^2}\sum_{\substack{i,j=1}}^n[{\log}(X)]_{i,j}
\end{aligned}
\end{equation}
Here, $\tilde{X}$ is the result of our proposed mean centralization operation applied to the SPD matrix $X$. \\	
	\indent Inspired by the broad applications of arc-cosine kernel \cite{cho2009kernel} in the Euclidean space and a family of Log-Euclidean kernels proposed in \cite{li2013log}, we propose Log-Euclidean framework based arc-cosine kernel(LogE.Arc kernel), which extends the well-known arc-cosine kernel onto the nonlinear Riemannian manifold of SPD matrices.\\
\begin{bfseries} Definition 5 \end{bfseries} (arc-cosine kernel)\emph{ Let $x,y$ $\in$ $R^d$ be two vectors. The arc-cosine kernel can be expressed as the angle $\theta$ between the samples [6] as:}
\begin{equation}
\label{eq9}
\mathop \theta = {\rm arccos}(\frac{x \cdot y}{\parallel x \parallel\parallel y \parallel})
\end{equation} 
\emph{In \cite{cho2009kernel}, the arc-cosine kernel has a simple formulation, which depends on the magnitude of the vectors and the angle between them. It can be defined as:}
\begin{equation}
\label{eq10}
\mathop {k{_r}}(x,y) =\frac{1}{\pi}{\parallel x \parallel}^r {\parallel y \parallel}^r J_r(\theta)
\end{equation}
\emph{where the angular dependence function $J_r$($\theta$) for different orders $r\ge0$ is defined as:}\\
\begin{equation}
\label{eq11}
\mathop {J{_r}}(\theta) = (-1)^r (\sin \theta){^{2r+1}}(\frac{1}{\sin \theta} \frac{\partial}{\partial\theta})^r (\frac{\pi - \theta}{\sin \theta})
\end{equation}
The arc-cosine kernel function  $k_r(x,y)$ has different properties that are shared respectively by radial basis function (RBF), linear, and polynomial kernels (Interested reader can refer to \cite{cho2009kernel} for the details of the arc-cosine kernel). Motivated by the work in \cite{li2013log} and the Log-Euclidean framework, the inputs, $x$, and $y$, of arc-cosine kernel can not only be vectors in the Euclidean space, but also SPD matrices on the curved Riemannian manifold. Thus, the Log-Euclidean framework based arc-cosine kernels can be defined as:
\begin{equation}
\begin{aligned}
\label{eq12}
\mathop {{k}{_{LogE.Arc}^r}}(x,y)=\frac{1}{\pi}{\parallel\log(x)\parallel} {_F^r}{\parallel\log(y)\parallel}{_F^r}J{_r}(\theta)
\end{aligned}
\end{equation} 
Here $x,y$ $\in$ $S_n^{++}$ and $J{_r}(\theta)$ has the same formulation as Eq.11, $\theta$ is the angle between the inputs that are mapped into the domain of logarithm matrix:
\begin{equation}
\begin{aligned}
\label{eq13}
\mathop \theta = {\rm arccos}(\frac{Tr(\log(x)\log(y))}{{\parallel\log(x)\parallel}{_F}\;{\parallel\log(y)\parallel}{_F}})
\end{aligned}
\end{equation} 
The  arc-cosine kernel given in (Eq.12) sets up a Log-Euclidean framework for constructing kernels on SPD manifold, which measures the similarity of SPD matrices and referred to as Log-Euclidean framework based arc-cosine kernel (LogE.Arc kernel). The LogE.Arc kernels of different orders are the corresponding arc-cosine kernels in the domain of logarithm matrix which inherit the corresponding property (RBF, etc.) in the vector space.

\begin{figure*}
\begin{center}
\includegraphics[width=0.85 \linewidth]{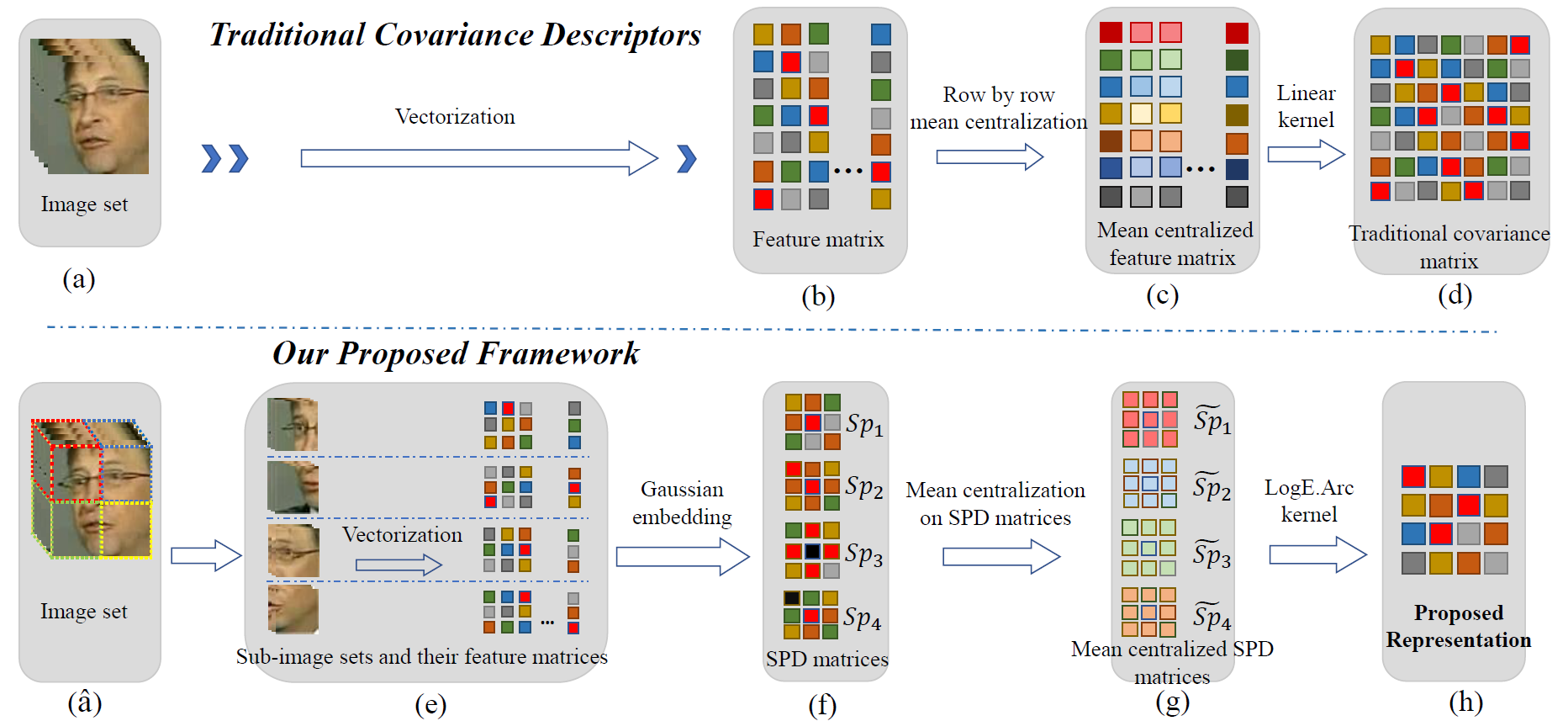} 
\end{center}
   \caption{ The flow chart of traditional covariance descriptors (CovDs) and our proposed framework (CovDs-S). Top: traditional CovDs for describing image set. Bottom: our proposed framework for describing image set.}
\label{fig:short}
\end{figure*}

\section{Proposed Framework}
	\indent In this section, we first give a brief overview of traditional CovDs for describing image sets and then present our proposed CovDs-S. Finally, we compare our CovDs-S with other improved versions of CovDs.

\subsection{Traditional Covariance Descriptors}
	\indent Consider a feature matrix (Fig.2 step (b)) of an image set with $n$ images: $S$ = $[s_1,s_2,...,s_n]$, where $s_i$ $\in$ $R^d$ is the $d$-dimensional feature vector representing the $i$-th image. Color images need to be processed as grayscale images and the $d$-dimensional feature vectors are obtained by vectorizing the grayscale images. Using the traditional CovDs, the representation \cite{wang2012covariance} of this image set can be obtained by:
\begin{equation}
\begin{aligned}
\label{eq14}
\mathop C = \frac{1}{n-1} \sum_{i=1}^n (s_i - \tilde{s}){(s_i - \tilde{s})}{^T} =  \tilde{S}{\tilde{S}}^T \\
where \qquad \tilde{S}=(n-1)^{-\frac{1}{2}}(S-\tilde{s}1_n^T) \qquad
\end{aligned}
\end{equation}
and $\tilde{s}$ = $\frac{1}{n}\sum_{i=1}^ns_i$ is a $d$-dimensional mean vector of the feature matrix $S$. $\tilde{S}$ is the mean centralized matrix (Fig.2 step (c)), and $1_n$ is a column vector of $n$ ones. The covariance matrix, $C$, can also be viewed as the kernel matrix between mean centralized feature vectors of the corresponding pixels (the rows of mean centralized matrix $\tilde{S}$) via linear kernel (Fig.2 step (d)):\\
\begin{equation}
\begin{aligned}
\label{eq15}
\mathop C = (C_{i,j}){_{i,j=1,...,d}} \qquad \qquad \quad \quad \\
where \quad  {C_{i,j}}={k}_{linear} (\tilde{S}_{(i,:)},\tilde{S}_{(j,:)}) = \tilde{S}_{(i,:)}\tilde{S}_{(j,:)}^T 
\end{aligned}
\end{equation}
where $\tilde{S}_{(i,:)}$ $\in$ $R^n$ denotes the $i$-th row of $\tilde{S}$, which is also the feature vector that represents $i$-th pixel of $n$ images. $C_{i,j}$ is the result of a linear kernel operation between $\tilde{S}_{(i,:)}$ and $\tilde{S}_{(j,:)}$ and denotes the similarity between $i$-th pixel and $j$-th pixel. 



\subsection{Proposed Framework for Image Set Coding}
	\indent Our proposed framework offers lower-dimensional and more discriminative representation for describing image sets than the traditional CovDs. For the sub-volumes of an image set (Fig.2 step ($\hat{a}$)), namely sub-image sets, we use SPD matrices to describe them via a Gaussian embedding. We then centralize these SPD matrices and use the LogE.Arc kernel to operate on the resulting mean centralized SPD matrices. Finally, the image set representation in our proposed framework is the kernel matrix defined by the mean centralized SPD matrices associated with the corresponding sub-image sets. We first give a brief overview of the Gaussian embedding and  elaborate the bottom row of Figure 2 to describe our framework.\\
	\indent The feature matrix $S = [s_1,s_2,...,s_n]$, as introduced in the sub-section of traditional CovDs, can also be described by a Gaussian model. The space spanned by a Gaussian model is a Riemannian manifold, and Gaussian embedding can embed this special manifold into SPD manifold \cite{wang2016raid}:
\begin{equation}
\label{eq16}
\mathop N(\mu,\Sigma) \sim G(\beta) = \left[  \begin{array}{cc}
      \Sigma + {\beta}^2{\mu}{\mu}^T & \beta\mu \\
      \beta{\mu}^T & 1
    \end{array} \right]
\end{equation}
where $\mu$ is a $d$-dimensional mean vector of $S$ and $\Sigma$ is a real $d$$\times$$d$ covariance matrix. $\beta>0$ is a parameter balancing the covariance matrix and the mean vector, The resulting matrix $G$($\beta$) $\in$ $R^{(d+1)\times(d+1)}$ is an SPD matrix and used as descriptors for sub-image sets(Fig.2 step (f)) in our proposed framework.\\
	\indent The bottom row of Figure 2 shows the flow chart of our proposed framework. We first obtain 4 sub-image sets via a sliding window(Fig.2 step ($\hat{a}$),\emph{ where the particular 4 sub-image were selected just for demonstration. Their choice is not fixed in our framework}). Then we use four SPD matrices (Fig.2 step (f)) to represent sub-image sets via Gaussian embedding (Eq.8) and obtain four mean centralized SPD matrices: $\tilde{Sp}_1$, $\tilde{Sp}_2$, $\tilde{Sp}_3$, $\tilde{Sp}_4$(Fig.2 step (g)) via the operation of mean centralization. Finally, the resulting representation (Fig.2 step (h)) is the sum kernel matrix of the four mean centralized SPD matrices via the LogE.Arc kernels of different orders(Eq.12). To this end, the resulting representation can generally be defined as:
\begin{equation}
\begin{aligned}
\label{eq17}
\mathop {{C}_{CovDs-S}} = \sum_{r=0}^R w_rC_r\qquad \qquad  \\
where \quad C_r = ([C_r]_{i,j})_{i,j=1,...,N} \quad \\
 and \quad {[C_r]_{i,j}} = k_{LogE.Arc}^r(\tilde{Sp}_i,\tilde{Sp}_j)
\end{aligned}
\end{equation}
where $N$ is the number of the sub-image sets, which is the key parameter determining the dimensionality, $R$ is the number of orders selected for LogE.Arc kernel. $\tilde{Sp}_i$ is the mean centralized SPD matrix, which is the representation of the $i$-th sub-image set. $C_r$ is the local kernel matrix  using the $r$-th order LogE.Arc kernel function between mean centralized SPD matrices:  $\tilde{Sp}_1$,$\tilde{Sp}_2$,...,$\tilde{Sp}_N$. The resulting representation: $C_{CovDs-S}$ is the sum kernel matrix (The sum of kernels is also a $p.d$ kernel \cite{sun2017learning}) of the local kernel matrices multiplied by the corresponding weight $w_r$. ${[C_{CovDs-S}]}_{i,j}$ denotes the similarity of the $i$-th sub-image set and $j$-th sub-image set. 

\subsection{Learning weights via Kernel Alignment}
	\indent In this sub-section, we learn, via the kernel target alignment, the weight coefficients $w_r$ associated with the $r$-th order LogE.Arc kernel for our proposed framework via the kernel target alignment.\\
	\begin{bfseries} Definition 6 \end{bfseries} (Kernel Alignment \cite{cortes2010two,sun2017learning}) \emph{The kernel alignment aims to align an input kernel matrix $K$ to a target kernel matrix $K_T$. It is defined as:}
\begin{equation}
\begin{aligned}
\label{eq18}
\mathop  \rho(K,K_T)=\frac{<K,K_T>_F}{\sqrt{<K,K>_F<K_T,K_T>_F}}
\end{aligned}
\end{equation}
the result of Eq.18 can be viewed as the cosine of the angle between $K$ and $K_T$. The weight coefficients $W=[w_0;...;w_R]$ should be estimated by maximizing the $\rho(K_W,K_T )$. $K_W$ is the global kernel matrix obtained as the sum of local kernel matrices of different orders. The kernel alignment has the following optimization formulation \cite{cortes2010two}:
\begin{equation}
\begin{aligned}
\label{eq19}
\mathop  {W^{*}} = {\rm arg\, max}\, \rho(K_W,K_T) = {\rm arg\, max}\, \frac{Tr(K_WK_T)}{\sqrt{Tr(K_WK_W)}}
\end{aligned}
\end{equation}
We now  introduce the kernel matrices: $K_W$ and $K_T$. Given a set of training samples $X=[x_1,x_2,...,x_N]$, where $x_i=\sum_{r=0}^R{w_r}{C_r^i}$ is our proposed representation for the $i$-th image set and $C_r^i$ is the $r$-th order LogE.Arc kernel matrix between sub-image sets constructed  from the $i$-th image set by image division, and the corresponding label matrix  $Y=[y_1,y_2,...,y_N]^T$ for the $N$ samples, where $y_i$ $\in$ $R^C$ contains the class label information of $i$-th sample and the $c$-th element of $y_i$ is 1 if $x_i$ is from the $c$-th class. In this paper, the global kernel matrix $K_W$ is the sum of $K_w^r$ multiplied by the weight $w_r$: $K_W=\sum_{r=0}^Rw_rK_w^r$ and  $K_w^r$ is the local kernel matrix between $C_r^1,...,C_r^N$: $[K_w^r]_{i,j}= Tr(C_r^iC_r^j)$. The target kernel matrix $K_T$ is defined via label matrix: $K_T=YY^T$. As introduced in \cite{cortes2010two}, the objective function in Eq.19 can be rewritten as:
\begin{equation}
\begin{aligned}
\label{eq20}
\mathop  {W^{*}} = \mathop{\rm arg\, max}_{\parallel W \parallel =1}\, \frac{W^T\beta{\beta}^TW}{W^T\Omega W} 
\end{aligned}
\end{equation}
where $\parallel W \parallel$  = 1 is a regularization term. $\beta$ is defined by $\beta_i$ = $Tr(\hat{K}_w^i K_T)$, where $\hat{K}_w^i$ is the centralized matrix of the kernel matrix $K_w^i$ \cite{cortes2010two,sun2017learning}, and the matrix $\Omega$ is defined by $\Omega_{i,j} = Tr(\hat{K}_w^i \hat{K}_w^j)$. \\
	\indent According to Proposition 2 in \cite{cortes2010two}, the solution $W^{*}$ of Eq.20 is given by:
\begin{equation}
\begin{aligned}
\label{eq21}
\mathop  {{W}^{*}} = \frac{\Omega^{-1}\beta}{\parallel \Omega^{-1}\beta \parallel}
\end{aligned}
\end{equation}

\subsection{Comparison with other Improved Versions of traditional CovDs for Image Set Coding}
	\indent As far as we know, the works in \cite{wang2015beyond,li2016spatial,chen2018component} are the only three improved versions of traditional CovDs for describing image sets. It is desirable to manifest their connections and differences.\\
	\indent Wang et al \cite{wang2015beyond} proposed an open framework\footnote{https://www.uow.edu.au/~leiw/} to use the kernel matrix over feature dimensions as a generic representation. This work uses a non-linear kernel matrix as the representation, but the kernel functions are defined in the Euclidean space and the resulting representation describes similarities between pixels at different locations, as the traditional CovDs \cite{wang2015beyond}. Our work proposes to capture the similarities between sub-image sets that contain more useful information. It extends the acr-cosine kernel onto the SPD manifold for this purpose. \\
	\indent Li et al \cite{li2016spatial} extended the descriptive granularity of covariance matrix from traditional pixel-level to more general patch-level. Though this work concentrates on the patch-level covariance computation, it is actually  a sum-pooling form of the pixel-level covariance. There is an essential difference in the way of descriptor computation. For describing image sets, we use the kernel matrix computed on  SPD manifold as the resulting representation instead of the covariance matrix computed in the Euclidean space \cite{li2016spatial}. Moreover, we use the SPD matrices to represent sub-image sets instead of the feature matrices consisting of intensity values \cite{li2016spatial}. \\
	\indent Chen et al \cite{chen2018component} proposed a framework\footnote{https://github.com/Kai-Xuan/ComponentSPD/} to generate low-dimensional discriminative data representation for describing image sets and concentrated on characterising the similarities between sub-image sets instead of pixels. The main difference his approach and our method is that we combine arc-cosine kernels of different orders in the domain of logarithm matrix instead of using a linear kernel \cite{chen2018component}. Moreover, we obtain our sub-image sets via the sliding window technique instead of dividing images into non-overlapping blocks \cite{chen2018component} and use the Gaussian embedding to model them. The work in \cite{chen2018component} is a special case of our proposed framework.

\begin{table*}
\renewcommand\arraystretch{1.12}
\centering
\caption{\label{tab:overlapping} \upshape Average recognition rates and standard deviations of different descriptors.}
\vspace {1.0mm}
\tiny   
\resizebox{5.4in}{!}{
\begin{tabular}{|c|c|c|c|c|c|c|c|c|c|}
\hline 
Methods 	&Descriptors	&CG\cite{kim2009canonical}   &ETH-80\cite{leibe2003analyzing}  &Virus\cite{kylberg2011virus}  &MDSD\cite{shroff2010moving}  \\
\hline
\multirow{5}*{NN-AIRM} 
&CovDs \cite{wang2012covariance}	&51.82$\pm$2.55			&70.12$\pm$5.24			&27.57$\pm$4.34 		&13.08$\pm$4.05\\
\cdashline{2-7}[0.5pt/1.5pt]
&CovDs-B \cite{wang2015beyond}		&71.87$\pm$2.47  		&89.17$\pm$3.48  		&36.57$\pm$5.00 		&23.67$\pm$5.55\\
\cdashline{2-7}[0.5pt/1.5pt]
&CovDs-P \cite{li2016spatial}		&87.49$\pm$1.41  		&87.83$\pm$4.61  		&67.40$\pm$6.19 		&22.51$\pm$4.56\\
\cdashline{2-7}[0.5pt/1.5pt]
&CovDs-C \cite{chen2018component}	&89.07$\pm$1.15  		&\textbf{94.53$\pm$2.55}  	&40.17$\pm$6.07 		&21.13$\pm$5.68\\
\cdashline{2-7}[0.5pt/1.5pt]
&CovDs-S (Ours)					&\textbf{90.21$\pm$1.27}  	&94.18$\pm$3.69  		&\textbf{67.60$\pm$4.76} 	&\textbf{32.38$\pm$5.60}\\
\hline
\multirow{5}*{NN-Stein} 
&CovDs \cite{wang2012covariance}	&40.66$\pm$2.62			&57.08$\pm$5.62			&27.43$\pm$4.90 		&12.67$\pm$4.25\\
\cdashline{2-7}[0.5pt/1.5pt]
&CovDs-B \cite{wang2015beyond}		&75.06$\pm$2.32  		&88.32$\pm$3.71  		&35.17$\pm$4.48 		&22.92$\pm$6.17\\
\cdashline{2-7}[0.5pt/1.5pt]
&CovDs-P \cite{li2016spatial}		&79.97$\pm$2.30  		&88.75$\pm$4.69  		&67.03$\pm$6.14 		&21.77$\pm$4.16\\
\cdashline{2-7}[0.5pt/1.5pt]
&CovDs-C \cite{chen2018component}	&89.76$\pm$1.23  		&94.10$\pm$2.90  		&39.83$\pm$6.18 		&19.51$\pm$5.42\\
\cdashline{2-7}[0.5pt/1.5pt]
&CovDs-S (Ours)					&\textbf{90.30$\pm$1.28}  	&\textbf{94.18$\pm$3.69}  	&\textbf{67.70$\pm$4.80} 	&\textbf{31.51$\pm$5.89}\\
\hline
\multirow{5}*{NN-Jeffrey} 
&CovDs \cite{wang2012covariance}	&82.45$\pm$1.38			&86.12$\pm$4.81			&30.80$\pm$5.49 		&18.56$\pm$4.77\\
\cdashline{2-7}[0.5pt/1.5pt]
&CovDs-B \cite{wang2015beyond}		&59.38$\pm$2.42  		&90.13$\pm$3.58  		&40.03$\pm$5.81 		&23.67$\pm$5.37\\
\cdashline{2-7}[0.5pt/1.5pt]
&CovDs-P \cite{li2016spatial}		&83.07$\pm$1.44  		&87.63$\pm$4.54  		&65.37$\pm$6.55 		&21.67$\pm$4.29\\
\cdashline{2-7}[0.5pt/1.5pt]
&CovDs-C \cite{chen2018component}	&89.52$\pm$1.14  		&94.60$\pm$2.79  		&40.97$\pm$6.54 		&24.00$\pm$5.30\\
\cdashline{2-7}[0.5pt/1.5pt]
&CovDs-S (Ours)					&\textbf{90.02$\pm$1.22}  	&\textbf{94.68$\pm$3.64}  	&\textbf{67.50$\pm$4.79} 	&\textbf{33.03$\pm$5.72}\\
\hline
\multirow{5}*{NN-LEM} 
&CovDs \cite{wang2012covariance}	&67.47$\pm$1.93			&78.17$\pm$5.59			&25.97$\pm$4.62 		&13.74$\pm$4.52\\
\cdashline{2-7}[0.5pt/1.5pt]
&CovDs-B \cite{wang2015beyond}		&73.18$\pm$2.39  		&90.65$\pm$3.79  		&36.07$\pm$4.22 		&24.87$\pm$5.51\\
\cdashline{2-7}[0.5pt/1.5pt]
&CovDs-P \cite{li2016spatial}		&89.44$\pm$1.17  		&88.27$\pm$4.24  		&67.57$\pm$6.61 		&24.90$\pm$4.96\\
\cdashline{2-7}[0.5pt/1.5pt]
&CovDs-C \cite{chen2018component}	&90.24$\pm$1.09  		&93.48$\pm$3.16  		&40.57$\pm$5.53 		&21.54$\pm$5.31\\
\cdashline{2-7}[0.5pt/1.5pt]
&CovDs-S (Ours)					&\textbf{90.49$\pm$1.31}  	&\textbf{94.07$\pm$3.77}  	&\textbf{68.10$\pm$4.88} 	&\textbf{32.23$\pm$6.08}\\
\hline
\multirow{5}*{Ker-SVM} 
&CovDs \cite{wang2012covariance}	&91.54$\pm$1.16			&92.60$\pm$5.15			&65.83$\pm$5.63 		&35.74$\pm$6.11\\
\cdashline{2-7}[0.5pt/1.5pt]
&CovDs-B \cite{wang2015beyond}		&92.31$\pm$1.13  		&94.18$\pm$3.69  		&73.77$\pm$5.82 		&37.79$\pm$5.76\\
\cdashline{2-7}[0.5pt/1.5pt]
&CovDs-P \cite{li2016spatial}		&94.34$\pm$1.02  		&94.52$\pm$3.64  		&75.40$\pm$6.01 		&35.54$\pm$6.47\\
\cdashline{2-7}[0.5pt/1.5pt]
&CovDs-C \cite{chen2018component}	&93.81$\pm$1.01  		&95.45$\pm$2.85  		&53.63$\pm$6.80 		&38.08$\pm$6.05\\
\cdashline{2-7}[0.5pt/1.5pt]
&CovDs-S (Ours)					&\textbf{94.36$\pm$0.98}  	&\textbf{97.07$\pm$2.66}  	&\textbf{77.93$\pm$5.03} 	&\textbf{43.92$\pm$6.09}\\
\hline
\end{tabular}}
\end{table*}

\section{Experiments and Analysis}
	\indent This section presents comparative experimental results of our proposed framework with state-of-the-art (SOTA) methods for the task of image set classification. 
\subsection{Datasets and settings}
	\indent In our first experiment involving the task of image set classification, we consider the Cambridge hand-gesture (CG) dataset \cite{kim2009canonical} that contains nine categories of samples and nine hundred image sets. Each class has twenty image sets chosen for training at random, and the remaining eighty image sets are reserved for testing. In the ETH-80 dataset \cite{leibe2003analyzing}, there are eight categories of samples and eighty image sets. For each class, five image sets are randomly chosen as training samples and the remaining five image sets are used for testing.  In the Virus cell dataset \cite{kylberg2011virus}, there are fifteen categories of samples and 100 images in each category. We divided the images of each category equally into five different image sets and obtained seventy-five image sets. For each class, three image sets are randomly chosen as training samples and the remaining two image sets for testing. The MDSD dataset \cite{shroff2010moving} has been used for the task of dynamic scene classification. Following the settings in \cite{sun2017learning}, we test the method based on the protocol of seventy-thirty-ratio (STR) which chooses seven videos for training and three videos for testing in each class. 

\begin{table*}[ht]
\renewcommand\arraystretch{1.15}
\centering
\caption{\label{tab:overlapping}Average recognition rates and standard deviations of different classifiers.}
\vspace {1.0mm}
\label{tab:1}
\tiny  
\resizebox{5.2in}{!}{
\begin{tabular}{cccccc}
\hline 
Methods		&CG\cite{kim2009canonical}  	&ETH-80\cite{leibe2003analyzing}    	&Virus\cite{kylberg2011virus} 		&MDSD\cite{shroff2010moving}\\
\hline
COV-LDA \cite{wang2012covariance}		&90.25$\pm$1.64	 	&93.95$\pm$4.30		&46.40$\pm$5.76		&34.10$\pm$5.90\\
COV-PLS \cite{wang2012covariance}		&88.95$\pm$1.26		&94.23$\pm$4.63		&62.84$\pm$5.99		&36.74$\pm$5.62\\
LogEKSR.Pol \cite{li2013log}			&92.32$\pm$1.19		&95.00$\pm$3.28		&58.53$\pm$6.54		&36.23$\pm$6.81\\
LogEKSR.Exp \cite{li2013log}			&92.23$\pm$1.18		&95.10$\pm$3.20		&59.03$\pm$6.38		&36.59$\pm$6.88\\
LogEKSR.Gau \cite{li2013log}			&92.33$\pm$1.18		&95.18$\pm$3.30		&61.80$\pm$6.35		&37.95$\pm$6.83\\
LEML	\cite{huang2015log}				&88.18$\pm$1.29		&93.05$\pm$3.31		&33.00$\pm$5.70		&25.97$\pm$6.79\\
LEML+COV-LDA \cite{huang2015log}		&89.09$\pm$1.63		&95.35$\pm$3.50		&58.03$\pm$5.84		&31.92$\pm$6.44\\
LEML+COV-PLS \cite{huang2015log}		&86.36$\pm$1.35		&95.83$\pm$3.04		&59.40$\pm$6.22		&35.90$\pm$6.98\\
SPDML-LEM \cite{harandi2018dimensionality}	&84.03$\pm$1.04		&90.63$\pm$4.19		&49.37$\pm$7.46		&24.23$\pm$4.47\\
SPDNet \cite{huang2017riemannian}		&92.03$\pm$1.46		&95.50$\pm$3.69		&59.70$\pm$4.58		&33.76$\pm$5.04\\
MMML \cite{wang2018multiple}			&92.87$\pm$1.39		&95.28$\pm$3.80		&51.13$\pm$7.60		&31.95$\pm$6.26\\
\hline
KS-CS-LEK			&93.63$\pm$1.08				&95.38$\pm$2.92				&74.93$\pm$5.92				&39.33$\pm$7.00\\
KS-CS-LogE.Pol			&93.95$\pm$0.94				&\textbf{97.30$\pm$2.55}		&75.17$\pm$4.86				&42.72$\pm$6.20\\
KS-CS-LogE.Exp			&93.68$\pm$0.90				&95.40$\pm$3.03				&70.90$\pm$5.60				&42.67$\pm$7.04\\
KS-CS-LogE.Gau			&93.90$\pm$0.91				&95.65$\pm$2.86				&71.87$\pm$5.18				&41.15$\pm$6.33\\
KS-CS-LogE.Arc 			&\textbf{94.36$\pm$0.98}		&97.07$\pm$2.66				&\textbf{77.93$\pm$5.03}		&\textbf{43.92$\pm$6.09}\\
\hline
\end{tabular}}
\end{table*}
\subsection{A Comparison with Existing Descriptors}
	\indent  For the comparative experiments with existing descriptors \cite{wang2015beyond,li2016spatial,chen2018component}, we first resize all images to $24\times 24$ and then use the intensity values to generate their corresponding representations. For our proposed framework, the sub-image sets are obtained by $6\times 6$ sliding window with spatial step of 2 pixels for the CG, ETH-80 and MDSD datasets, and spatial step of 3 pixels for the Virus dataset. In total, we obtain 100 sub-image sets for theCG, ETH-80 and MDSD datasets and 49 sub-image sets for the Virus dataset. In our framework, we set parameter $r$ in LogE.Arc kernel to be $r$ = [0, 1, 2, 3], and the value of $\beta$ in Eq.16 depends on the datasets, which is 0.05, 0.9, 14, 2 for the four datasets respectively. For the learned $W=[w_0,...,w_R]$, we set first two largest absolute values to be 1 and another two values to be zero on ETH-80, Virus and MDSD datasets. We set the largest absolute value to be 1 and another three values to be zero on theCG datasets. We regularize the traditional CovDs: $C^*=C+\lambda I$ to avoid the matrix singularity as introduced in \cite{wang2012covariance}, and set $\lambda$ to $10^{-3}\times Tr(C)$. To generate the descriptor in \cite{wang2015beyond}, we obtain the final kernel matrix representation via RBF kernel that has been shown to produce better accuracies \cite{wang2015beyond}. For the fairness of the comparative experiments, the patch size in \cite{li2016spatial} and block size in \cite{li2016spatial} are all $6\times 6$ and the step size is the same as the setting used by our proposed CovDs-S on the corresponding dataset. The different descriptors evaluated in our experiments are referred to as: \\
\begin{bfseries} CovDs:\end{bfseries} Image set repesented by traditional CovDs\cite{wang2012covariance}.
\begin{bfseries} CovDs-B:\end{bfseries} Image set repesented by the method in \cite{wang2015beyond}.
\begin{bfseries} CovDs-P:\end{bfseries} Image set repesented by the method in \cite{li2016spatial}.
\begin{bfseries} CovDs-C:\end{bfseries} Image set repesented by the method in \cite{chen2018component}.
\begin{bfseries} CovDs-S:\end{bfseries} Image set repesented by our framework.\\
	\indent In our experiments, five classification algorithms are used to verify the validity of our proposed CovDs-S, which include four nearest neighbor (NN) algorithms based on AIRM, Stein divergence, Jeffrey divergence, LEM and the well-known SVM classifier \cite{chang2011libsvm}. The different methods tested in our experiments are referred to as: \\
\begin{bfseries} NN-AIRM:\end{bfseries} AIRM-based NN classifier.\\
\begin{bfseries} NN-Stein:\end{bfseries} Stein divergence-based NN classifier.\\
\begin{bfseries} NN-Jeffrey:\end{bfseries} Jeffrey divergence-based NN classifier.\\
\begin{bfseries} NN-LEM:\end{bfseries} LEM-based NN classifier.\\
\begin{bfseries} Ker-SVM:\end{bfseries} LEK-based SVM classifier.\\
	\indent Here, Ker-SVM is a one-vs-all SVM classifier\footnote{http://www.peihuali.org/publications/RAID-G/RIAD-G\_V1.zip} implemented by Wang et al. Table 1 shows the average recognition rates and standard deviations of different descriptors with the same classifiers. In addition to the results with NN-AIRM on the ETH-80 dataset, the recognition rates of CovDs-S are higher than other four descriptors while using the same classification algorithm. This confirms that our CovDs-S captures more discriminative information than the other methods. In particular, the accuracy of CovDs-C is not as high as shown in Table 1 if the setting follows the recommendations made in \cite{chen2018component}.

\subsection{Comparison with Existing Classifiers}
 	\indent Here, we compare our method with the SOTA algorithms including Covariance Discriminative Learning (COV-LDA, COV-PLS) \cite{wang2012covariance}, Log-Euclidean Kernels for Sparse Representation (LogEKSR.Pol, LogEKSR.Exp and LogEKSR.Gau) \cite{li2013log}, SPD Manifold Learning based on LEM (SPDML-LEM) \cite{harandi2018dimensionality}, Log-Euclidean Metric Learning (LEML, LEML+COV-LDA, LEML+COV-PLS) \cite{huang2015log}, Riemannian Network for SPD Matrix Learning (SPDNet) \cite{huang2017riemannian} and Multiple Manifolds Metric Learning (MMML) \cite{wang2018multiple}. For these methods, we first resize all images to $20\times20$ and use the intensity values as their features. \\
	\indent We use the Ker-SVM tested on the representations obtained by our framework as our proposed classification algorithm. In addition to the LogE.Arc kernel (introduced above) used in our framework, we also consider other types of kernel functions to enrich our framework, such as Log-Euclidean Kernel (LEK) \cite{wang2012covariance}, Log-Euclidean based polynomial (LogE.Pol), exponential (LogE.Exp) and Gaussian kernels (LogE.Gau). Our framework based classifiers are reffered as:\\
\begin{bfseries} KS-CS-LogE.Arc:\end{bfseries} Ker-SVM tested on CovDs-S (introduced above).\\
\begin{bfseries} KS-CS-LEK:\end{bfseries} Ker-SVM tested on the representations via our framework where the LogE.Arc kernel replaced by LEK\\
\begin{bfseries} KS-CS-LogE.Pol:\end{bfseries} Ker-SVM tested on the representations via our framework where the LogE.Arc kernelis replaced by LogE.Pol kernel\\
\begin{bfseries} KS-CS-LogE.Exp:\end{bfseries} Ker-SVM tested on the representations obtained by our framework, where the LogE.Arc kernel is replaced by LogE.Exp kernel\\
\begin{bfseries} KS-CS-LogE.Gau:\end{bfseries} Ker-SVM tested on the representations produced by our framework, where the LogE.Arc kernel is replaced by LogE.Gau kernel\\
	\indent As shown in Table 2, the classifiers based on our framwork produce better performance than other SOTA methods. The advantage of our methods is very obvious on the Virus and MDSD datasets, where image samples contain a large amount of noise. Our method KS-CS-LogE.Arc achieves the best recognition rate of $94.36\%$, $77.93\%$ and $43.92\%$ on the CG, Virus and MDSD datasets. Our framework based KS-CS-LogE.Pol achieves the best recognition rate of $97.30\%$ on the ETH-80 dataset and KS-CS-LogE.Arc achieves the second best recognition rate of $97.07\%$.

\subsection{Ablation Experiments}
	\indent In this subsection, we validate the contributions of each component in CovDs-S and analyze the effect of image rotation and sub-image sizes. To this end, CovDs-S is used in the following variants: CovDs-S without Gaussian embedding (CovDs-S-GE),  CovDs-S without kernel alignment (Coves-S-KA), CovDs-S without mean centralization (CovDs-S-MC), CovDs-S with image rotation $90^\circ$ (CovDs-S-IM90),  CovDs-S with image rotation $180^\circ$ (CovDs-S-IM180) and CovDs-S with image rotation $270^\circ$ (CovDs-S-IM270).\\
	\indent Table 3 shows the accuracies and the standard deviations of the different kernel variants in conjunction with the Ker-SVM classifier on the four datasets. From the results in this table, we can conclude that the contribution of Gaussian embedding and kernel alignment are more significant than the effect of mean centralization of our CovDs-S. In theory, the mean centralization operation corresponds to the standard normalization operation used in traditional CovDs. It appears that in the case of our framework, this operation does not impact on accuracy. According to the last three rows of the table, CovDs-S is also not sensitive to image rotation.\\
	\indent Figure 3 shows the effect of sub-image size and step size on the recognition rates of KS-CS-LogE.Arc. In the  horizontal ordinate in the form of 'a/b', 'a' represents the sliding window size, and 'b' denotes step size. The results show clearly that the proposed framework performs bettter when the sub-image size is $6\times6$. In that case, our CovDs-S achieves the best accuracies on the CG, ETH-80 and MDSD datasets when the step size is 2 pixels. When the step size is set as 3 pixels, our CovDs-S achieves the best accuracies on the Viurs dataset.

\begin{table}
\renewcommand\arraystretch{1.15}
\centering
\caption{\label{tab:overlapping}The accuracies and standard deviations of variants.}
\vspace {1.0mm}
\label{tab:1}
\large
\resizebox{3.2in}{!}{
\begin{tabular}{cccccc}
\hline 
Descriptors &CG\cite{kim2009canonical}  	&ETH-80\cite{leibe2003analyzing}    	&Virus\cite{kylberg2011virus} 		&MDSD\cite{shroff2010moving}\\
\hline
CovDs-S-GE		&94.18$\pm$0.88	 	&96.73$\pm$2.90		&60.00$\pm$6.21		&40.10$\pm$6.68\\
CovDs-S-KA		&91.76$\pm$1.26		&96.62$\pm$3.02		&74.63$\pm$5.14		&43.79$\pm$6.16\\
CovDs-S-MC		&94.34$\pm$0.97		&97.07$\pm$2.66		&77.67$\pm$4.68		&43.74$\pm$6.10\\
CovDs-S-IM90	&94.32$\pm$1.00		&97.07$\pm$2.66		&77.40$\pm$4.70		&43.90$\pm$6.16\\
CovDs-S-IM180	&94.34$\pm$0.97		&97.07$\pm$2.66		&77.83$\pm$4.95		&43.92$\pm$6.06\\
CovDs-S-IM270	&94.33$\pm$0.99		&97.07$\pm$2.66		&77.53$\pm$4.56		&43.95$\pm$6.26\\
\hline
\end{tabular}}
\end{table}
\begin{figure}
\begin{center}
\includegraphics[width=0.99\linewidth]{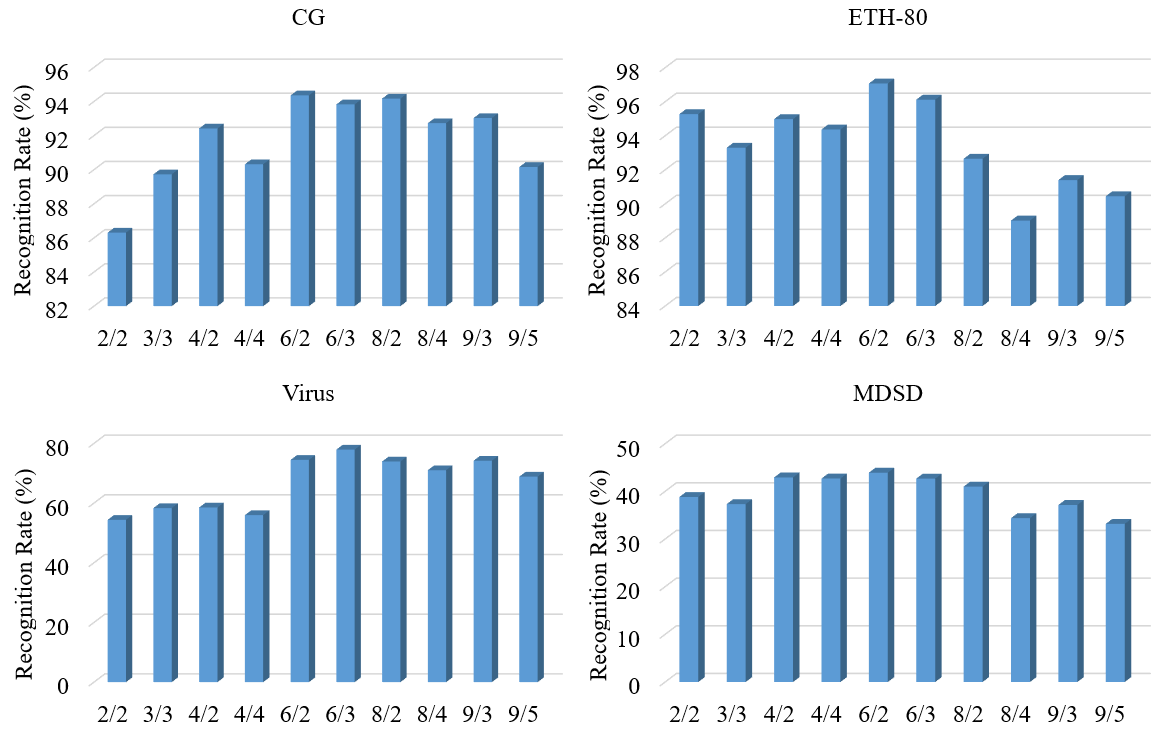} 
\end{center}
   \caption{ The infulence of different sub-image size and step size.}
\label{fig:short}
\end{figure}

\subsection{Advantages of Our Framework}
	\indent From the superior results in Table 1 and Table 2, we can conclude that our proposed framework captures more discriminative information for the task of image set classification than other methods. The superior performance is particularly notable for noisy samples. Moreover, the dimensionality of the representation obtained via our framework is related to the number of sub-image sets(100 or 49), which is far lower than that of the traditional CovDs. Table 4 shows the run time for representative methods (NN-AIRM and Ker-SVM), as well as the generating time (GT) required for different descriptors on the ETH-80 dataset, where the unit of time is one second. The representations extracted by our framework require far less time than  the traditional CovDs.  With the recommended settings in the experiments, our proposed representations tend to be nonsingular. The dimensionality of the feature representation for the sub-image set is $(1+(6^2+1))(6^2+1)/2=703$. The resulting representation can also be viewed as corresponding to the kernel matrix (arc-cosine kernel, etc. in the Euclidean space) of $100\times703$ or $49\times703$ feature matrix $S$ as traditional CovDs. Here, we cannot be sure about the nonsingularity for the resulting kernel matrix,  but it is more likely to hold for an SPD matrix than a traditonal covariance matrix. 
\begin{table}
\renewcommand\arraystretch{1.2}
\centering
\caption{\label{tab:overlapping} \upshape Time comparisons on ETH-80 dataset.}
\vspace {1.0mm}
\large
\resizebox{3.2in}{!}{
\begin{tabular}{|c|c|c|c|c|c|c|c|c|c|}
\hline
&\multirow{2}*{CovDs} 
&\multicolumn{5}{c|}{CovDs-S}\\
\cline{3-7}
	& 		&LEK		&LogE.Pol  	&LogE.Exp  	&LogE.Gau		&LogE.Arc\\
\hline
GT 		&20.15	&48.67	&48.74 		&49.32 		&49.20 		&51.83\\
\hline
NN-AIRM 	&28.77	&1.248	&1.252		&1.248		&1.250 		&1.249\\
\hline
Ker-SVM 	&2.07		&0.056	&0.056		&0.056 		&0.056 		&0.055\\
\hline
\end{tabular}}
\end{table}
\section{Conclusion And Future Work}
	\indent We proposed a novel framework extending CovDs from the Euclidean to an SPD manifold. It  generates a kernel matrix defined by the representations of sub-image sets instead of pixels. Our method provides a lower-dimensional data representation, which is beneficial for improving the efficiency of classifiers. The experimental results show that the representation obtained by our proposed framework is more discriminative  than other methods when performing  the task of image set classification. In future, we will consider how to extend our proposed framework to Reproducing Kernel Hilbert Space (RKHS).
\section{Acknowledgments}
	\indent THE PAPER IS SUPPORTED BY THE NATIONAL NATURAL SCIENCE FOUNDATION OF CHINA (GRANT NO 61672265,U1836218), THE 111 PROJECT OF MINISTRY OF EDUCATION OF CHINA (GRANT NO. B12018), UK EPSRC GRANT EP/N007743/1, AND MURI/EPSRC/DSTL GRANT EP/R018456/1.

{\small
\bibliographystyle{ieee}
\bibliography{egbib}

\begin{thebibliography}{10}\itemsep=-1pt

\bibitem{arandjelovic2013all}
R.~Arandjelovic and A.~Zisserman.
\newblock All about vlad.
\newblock In {\em Proceedings of the IEEE conference on Computer Vision and
  Pattern Recognition}, pages 1578--1585, 2013.

\bibitem{arsigny2007geometric}
V.~Arsigny, P.~Fillard, X.~Pennec, and N.~Ayache.
\newblock Geometric means in a novel vector space structure on symmetric
  positive-definite matrices.
\newblock {\em SIAM journal on matrix analysis and applications},
  29(1):328--347, 2007.

\bibitem{chang2011libsvm}
C.-C. Chang and C.-J. Lin.
\newblock Libsvm: a library for support vector machines.
\newblock {\em ACM transactions on intelligent systems and technology (TIST)},
  2(3):27, 2011.

\bibitem{chen2018component}
K.-X. Chen and X.-J. Wu.
\newblock Component spd matrices: A low-dimensional discriminative data
  descriptor for image set classification.
\newblock {\em Computational Visual Media}, 4(3):245--252, 2018.

\bibitem{chen2018riemannian}
K.-X. Chen, X.-J. Wu, R.~Wang, and J.~Kittler.
\newblock Riemannian kernel based nystr{\"o}m method for approximate
  infinite-dimensional covariance descriptors with application to image set
  classification.
\newblock In {\em 2018 24th International Conference on Pattern Recognition
  (ICPR)}, pages 651--656. IEEE, 2018.

\bibitem{cho2009kernel}
Y.~Cho and L.~K. Saul.
\newblock Kernel methods for deep learning.
\newblock In {\em Advances in neural information processing systems}, pages
  342--350, 2009.

\bibitem{cortes2010two}
C.~Cortes, M.~Mohri, and A.~Rostamizadeh.
\newblock Two-stage learning kernel algorithms.
\newblock In {\em ICML}, pages 239--246, 2010.

\bibitem{feng2017object}
F.~Feng, X.-J. Wu, and T.~Xu.
\newblock Object tracking with kernel correlation filters based on mean shift.
\newblock In {\em Smart Cities Conference (ISC2), 2017 International}, pages
  1--7. IEEE, 2017.

\bibitem{hamm2009extended}
J.~Hamm and D.~D. Lee.
\newblock Extended grassmann kernels for subspace-based learning.
\newblock In {\em Advances in neural information processing systems}, pages
  601--608, 2009.

\bibitem{harandi2018dimensionality}
M.~Harandi, M.~Salzmann, and R.~Hartley.
\newblock Dimensionality reduction on spd manifolds: The emergence of
  geometry-aware methods.
\newblock {\em IEEE transactions on pattern analysis and machine intelligence},
  40(1):48--62, 2018.

\bibitem{harandi2014bregman}
M.~Harandi, M.~Salzmann, and F.~Porikli.
\newblock Bregman divergences for infinite dimensional covariance matrices.
\newblock In {\em Proceedings of the IEEE Conference on Computer Vision and
  Pattern Recognition}, pages 1003--1010, 2014.

\bibitem{huang2017riemannian}
Z.~Huang and L.~Van~Gool.
\newblock A riemannian network for spd matrix learning.
\newblock In {\em Thirty-First AAAI Conference on Artificial Intelligence},
  2017.

\bibitem{huang2015log}
Z.~Huang, R.~Wang, S.~Shan, X.~Li, and X.~Chen.
\newblock Log-euclidean metric learning on symmetric positive definite manifold
  with application to image set classification.
\newblock In {\em International conference on machine learning}, pages
  720--729, 2015.

\bibitem{jegou2012aggregating}
H.~Jegou, F.~Perronnin, M.~Douze, J.~S{\'a}nchez, P.~Perez, and C.~Schmid.
\newblock Aggregating local image descriptors into compact codes.
\newblock {\em IEEE transactions on pattern analysis and machine intelligence},
  34(9):1704--1716, 2012.

\bibitem{kim2009canonical}
T.-K. Kim and R.~Cipolla.
\newblock Canonical correlation analysis of video volume tensors for action
  categorization and detection.
\newblock {\em IEEE Transactions on Pattern Analysis and Machine Intelligence},
  31(8):1415--1428, 2009.

\bibitem{kulis2009low}
B.~Kulis, M.~A. Sustik, and I.~S. Dhillon.
\newblock Low-rank kernel learning with bregman matrix divergences.
\newblock {\em Journal of Machine Learning Research}, 10(Feb):341--376, 2009.

\bibitem{kylberg2011virus}
G.~Kylberg, M.~Uppstr{\"o}m, and I.-M. Sintorn.
\newblock Virus texture analysis using local binary patterns and radial density
  profiles.
\newblock In {\em Iberoamerican Congress on Pattern Recognition}, pages
  573--580. Springer, 2011.

\bibitem{leibe2003analyzing}
B.~Leibe and B.~Schiele.
\newblock Analyzing appearance and contour based methods for object
  categorization.
\newblock In {\em Computer Vision and Pattern Recognition, 2003. Proceedings.
  2003 IEEE Computer Society Conference on}, volume~2, pages II--409. IEEE,
  2003.

\bibitem{li2013log}
P.~Li, Q.~Wang, W.~Zuo, and L.~Zhang.
\newblock Log-euclidean kernels for sparse representation and dictionary
  learning.
\newblock In {\em Proceedings of the IEEE International Conference on Computer
  Vision}, pages 1601--1608, 2013.

\bibitem{li2016spatial}
Y.~Li, R.~Wang, Z.~Cui, S.~Shan, and X.~Chen.
\newblock Spatial pyramid covariance-based compact video code for robust face
  retrieval in tv-series.
\newblock {\em IEEE Transactions on Image Processing}, 25(12):5905--5919, 2016.

\bibitem{peng2014action}
X.~Peng, C.~Zou, Y.~Qiao, and Q.~Peng.
\newblock Action recognition with stacked fisher vectors.
\newblock In {\em European Conference on Computer Vision}, pages 581--595.
  Springer, 2014.

\bibitem{pennec2006riemannian}
X.~Pennec, P.~Fillard, and N.~Ayache.
\newblock A riemannian framework for tensor computing.
\newblock {\em International Journal of computer vision}, 66(1):41--66, 2006.

\bibitem{perronnin2010improving}
F.~Perronnin, J.~S{\'a}nchez, and T.~Mensink.
\newblock Improving the fisher kernel for large-scale image classification.
\newblock In {\em European conference on computer vision}, pages 143--156.
  Springer, 2010.

\bibitem{shroff2010moving}
N.~Shroff, P.~Turaga, and R.~Chellappa.
\newblock Moving vistas: Exploiting motion for describing scenes.
\newblock In {\em Computer Vision and Pattern Recognition (CVPR), 2010 IEEE
  Conference on}, pages 1911--1918. IEEE, 2010.

\bibitem{sivic2003video}
J.~Sivic and A.~Zisserman.
\newblock Video google: A text retrieval approach to object matching in videos.
\newblock In {\em null}, page 1470. IEEE, 2003.

\bibitem{sra2012new}
S.~Sra.
\newblock A new metric on the manifold of kernel matrices with application to
  matrix geometric means.
\newblock In {\em Advances in neural information processing systems}, pages
  144--152, 2012.

\bibitem{sun2017learning}
H.~Sun, X.~Zhen, Y.~Zheng, G.~Yang, Y.~Yin, and S.~Li.
\newblock Learning deep match kernels for image-set classification.
\newblock In {\em Proceedings of the IEEE Conference on Computer Vision and
  Pattern Recognition}, pages 3307--3316, 2017.

\bibitem{tan2015grassmann}
H.~Tan, Z.~Ma, S.~Zhang, Z.~Zhan, B.~Zhang, and C.~Zhang.
\newblock Grassmann manifold for nearest points image set classification.
\newblock {\em Pattern Recognition Letters}, 68:190--196, 2015.

\bibitem{tuzel2008pedestrian}
O.~Tuzel, F.~Porikli, and P.~Meer.
\newblock Pedestrian detection via classification on riemannian manifolds.
\newblock {\em IEEE Transactions on Pattern Analysis \&amp; Machine
  Intelligence}, (10):1713--1727, 2008.

\bibitem{wang2015beyond}
L.~Wang, J.~Zhang, L.~Zhou, C.~Tang, and W.~Li.
\newblock Beyond covariance: Feature representation with nonlinear kernel
  matrices.
\newblock In {\em Proceedings of the IEEE International Conference on Computer
  Vision}, pages 4570--4578, 2015.

\bibitem{wang2016raid}
Q.~Wang, P.~Li, W.~Zuo, and L.~Zhang.
\newblock Raid-g: Robust estimation of approximate infinite dimensional
  gaussian with application to material recognition.
\newblock In {\em Proceedings of the IEEE Conference on Computer Vision and
  Pattern Recognition}, pages 4433--4441, 2016.

\bibitem{wang2012covariance}
R.~Wang, H.~Guo, L.~S. Davis, and Q.~Dai.
\newblock Covariance discriminative learning: A natural and efficient approach
  to image set classification.
\newblock In {\em Computer Vision and Pattern Recognition (CVPR), 2012 IEEE
  Conference on}, pages 2496--2503. IEEE, 2012.

\bibitem{wang2018multiple}
R.~Wang, X.-J. Wu, K.-X. Chen, and J.~Kittler.
\newblock Multiple manifolds metric learning with application to image set
  classification.
\newblock In {\em 2018 24th International Conference on Pattern Recognition
  (ICPR)}, pages 627--632. IEEE, 2018.

\end{thebibliography}
}

\end{document}